\documentclass{article}
\usepackage{spconf,amsmath,graphicx}

\usepackage{color}
\usepackage{enumitem}

\newcommand{\rg}[1]{{\bf [\textcolor{blue}{ROBERTO #1}{\bf ]}}}

\newcommand{\COMMENT}[1]{}

\usepackage{multirow}

\title{AUTOMATIC ASSESSMENT OF SPOKEN LANGUAGE PROFICIENCY \\OF NON-NATIVE CHILDREN}

\twoauthors
  {Roberto Gretter$^1$, Marco Matassoni$^1$, Katharina Allgaier$^{1,2}$, Svetlana Tchistiakova$^{1,3,4}$\sthanks{Erasmus Mundus funded joint degree in Language and Communication Technologies}, Daniele Falavigna$^1$}
	{{\small {(1) Fondazione Bruno Kessler, Trento (Italy), (2) University of Heidelberg (Germany)}}\\
	\small {(3) University of Trento (Italy), (4) University of Saarland (Germany)}\\
	}
  {C. Author-three, D. Author-four\sthanks{The fourth author performed the work
	while at ...}}
	{School C-D\\
	Department C-D\\
	Address C-D}

\begin{document}
\ninept
\maketitle
\begin{abstract}
This paper describes technology developed to automatically grade Italian students (ages 9-16) on their English and German spoken language proficiency. The students' spoken answers are first transcribed by an automatic speech recognition (ASR) system and then scored using a feedforward neural network (NN) that processes features extracted from the automatic transcriptions. In-domain acoustic models, employing deep neural networks (DNNs), are derived by adapting the parameters of an original out of domain DNN.
\COMMENT{
For ASR, difficulties include: {\it a)} recognition of child speech, {\it b)} recognition of non-native speech, {\it c)} insertion of spontaneous speech phenomena (hesitations, false starts, etc), {\it d)} presence of code switching in the response, i.e.\   multiple languages (English / German / Italian are used by students), {\it e)} presence of a significant level of background noise from a microphone that remains open for a fixed time interval, {\it f)} non-collaborative speakers (students often joke, laugh, speak softly, etc).
} 
Automatic scores are computed for  low level proficiency indicators - such as: lexical richness, syntax correctness, quality of pronunciation, discourse fluency, semantic relevance to the prompt, etc - defined by human experts in language proficiency.
A set of experiments was carried out on a large set of data collected during proficiency evaluation campaigns involving thousands of students, manually scored by human experts. Obtained results are presented and discussed.

\end{abstract}
\section{Introduction}
\label{sec:introduction}

The problem of automatic scoring of second language (L2) learning proficiency has been largely investigated in the past in the framework of computer assisted language learning (CALL). Approaches have been  proposed for two input modalities: written and spoken. In both cases, 
specific competencies of the human learners are processed by some suitable proficiency classifiers. The final goal is to measure L2 proficiency according to some standard scale. A well-known scale is the Common European Framework of Reference for Languages (Council of Europe, 2001). The CEFR defines $6$ levels of proficiency: A1 (beginner), A2, B1, B2, C1 and C2.

This work\footnote{This work has been partially funded by IPRASE (http://www.iprase.tn.it) under the project ``TLT -  Trentino Language Testing 2018''. We thank ISIT (http://www.isit.tn.it) for having provided the reference scores.}
addresses automatic scoring of L2 learners, focusing on different linguistic competences,  or ``indicators,'' related both to the content (e.g.\ grammatical correctness, lexical richness, semantic coherence, etc)  and to the speaking capabilities (e.g.\ pronunciation, fluency, etc).  Refer to Section~\ref{sec:speechdata} for a description of the  indicators adopted in this work. 
The learners are Italian students, between 9 and 16 years old, who study both English and German at school. The students took proficiency tests by answering question prompts provided in written form. Responses included typed answers and spoken answers. The developed system is based on a set of features (see Section \ref{sec:features}) 
extracted both from the input speech signal  and from the automatic transcriptions of the spoken responses. 
The features are classified with feedforward neural networks, trained on labels provided by human raters, who manually scored the ``indicators'' of 
a set of about $29,000$ spoken utterances (about $2,550$ proficiency tests).
Training and test data used in the experiments will be described in Section \ref{sec:speechdata}.

The task is very challenging and poses many problems which are only partially considered in the scientific literature. From the ASR perspective major difficulties are represented by: {\it a)} recognition of both child and non-native speech, i.e.\ Italian pupils speaking both English and German, {\it b)} presence of a large number of spontaneous speech phenomena (hesitations, false starts, fragments of words, etc.), {\it c)} presence of multiple languages (English, Italian and German words are frequently uttered in response to a single question), {\it d)} presence of a significant level of background noise due to the fact that the microphone remains opened for a fixed time interval (e.g.\ 20 seconds), and {\it e)} presence of non-collaborative speakers (students often joke, laugh, speak softly, etc.).

 

{\bf Relation to prior work.} Scientific literature is rich in approaches for automated assessment of spoken language proficiency. Performance is directly dependent on  ASR accuracy which, in turn, depends on the type of input, read or spontaneous, and on the speaker ages, adults or children (see  \cite{eskenazi2009} for an overview of spoken language technology for education).

Automatic assessment of reading capabilities of L2 children was widely investigated in the past at both sentence level \cite{zechsab2009}
 and word level \cite{tepper2007}.
More recently, the scientific community started addressing automatic assessment of more complex spoken tasks, requiring  more general communication capabilities by L2 learners.  The AZELLA data set  \cite{cheng2014}, developed by Pearson,  includes $1,500$ spoken tests, each double graded by human professionals, from a variety of tasks. The work  in  \cite{angeliki2014} describes a latent semantic analysis (LSA) based approach for scoring the proficiency of the AZELLA test set, while \cite{evanini2013} describes a system designed to automatically evaluate the communication skills of young English students.
Features proposed for evaluation of 
pronunciation are described for instance in 
\cite{kibishi2015statistical}.


Automatic scoring of  L2 proficiency has also been investigated in recent shared tasks. One of these \cite{baur2017} addressed a prompt-response task, where Swiss students learning English  had to answer to both written and spoken prompts. The goal is to label student spoken responses  as ``accept''  or ``reject''.  
The winners of the shared task \cite{oh2017} use a deep neural network (DNN) model to accept or reject input utterances, while the work reported in   \cite{evanini2017}  makes use of a support vector machine originally designed for scoring written texts.

Finally, it is worth mentioning that the recent end-to-end approach \cite{chen2018} (based on the usage of  a bidirectional recurrent DNNs  employing an attention model) performs  better than the well known SpeechRater\texttrademark  system \cite{zechner2009}, developed by ETS for automatically scoring non-native spontaneous speech in the context of an online practice test for prospective takers of the Test Of English as a Foreign Language (TOEFL)\footnote{TOEFL:  https://www.ets.org/toefl}.

With respect to the previously mentioned works, the novelties proposed in this paper are as follows. Firstly, we introduce a unique multi-lingual DNN, for acoustic modeling (AM), trained on English, German and Italian speech from children and adults. This model copes with {\em multi-lingual} spoken answers, i.e.\  utterances where the student uses or confuses words belonging to the three languages at hand. A common phonetic lexicon, defined in terms of units of the International Phonetic Alphabet (IPA), is adopted for transcribing the words of all languages. Moreover, spontaneous ``in-domain'' speech data (details in Section \ref{sec:ASR}) are included in the training material to model frequently occurring speech phenomena (e.g., laughs, hesitations, background noise).
We also propose a novel method to compute acoustic features using the phonetic representation and likelihoods output by the ASR system. We employ both our best non-native ASR system, and ASR systems trained only on native English/German data to generate these features (see subsection \ref{sec:features}).

Experimental results reported in the paper (see Section \ref{sec:experiments}) show that: {\it a)} the usage of the multilingual DNN is effective for transcribing non-native children's speech, {\it b)} the usage of feedforward NNs allows us to classify  each  indicator with an average classification accuracy between around $60$\% and around $67$\%, and {\it c)} no large differences in classification performance have been observed among the different indicators (i.e.\ the set of adopted features performs pretty well for all indicators). 

\vspace{-0.2cm}
\section{Description of the Data}
\label{sec:speechdata}

\subsection{Evaluation campaigns on trilinguism}

In Trentino (Northern Italy), a series of campaigns is underway for testing linguistic competencies of multilingual Italian students taking proficiency tests in English and German. 
A set of three evaluation campaigns were planned, taking place in 2016, 2017/2018, and 2020. Each one involves about 3000 students (ages 9-16), belonging to 4 different school grade levels, and three proficiency levels (A1, A2, B1). The 2017/2018 campaign was split into a group of 500 students in 2017, and 2500 students in 2018. Table~\ref{tab:plan} highlights some information about the 2018 campaign. Several tests aimed at assessing the language learning capabilities of the students were carried out by means of multiple-choice questions, which can be evaluated automatically. However, a detailed linguistic evaluation cannot be performed without allowing the students to express themselves in both written sentences and spoken utterances, which typically require the intervention of human experts to be scored. In this paper we will focus only on the spoken part of these proficiency tests.

\begin{table}[htb]
\caption{Evaluation of L2 linguistic competences in Trentino in 2018: level, grade, age and number of pupils participating in the English (ENG) and German (GER) tests.}
\label{tab:plan}
\centering
\begin{tabular}{ l c c r r r }
\\ \hline
CEFR  & Grade, School   & Age    & \#Pupils &\#ENG & \#GER  \\ \hline
 A1 & 5, primary        &  9-10  &  508 &  476 &  472 \\ 
 A2 & 8, secondary      & 12-13  &  593 &  522 &  547 \\ 
 B1 & 10, high school   & 14-15  & 1086 & 1023 &  984 \\ 
 B1 & 11, high school   & 15-16  &  364 &  310 &   54 \\ \hline
 \multicolumn{2}{c}{tot}&  9-16  & 2551 & 2331 & 2057 \\ \hline
\end{tabular}
\end{table}



Table~\ref{tab:dataspeech} reports some statistics extracted from  the spoken data collected in 2017/2018. 
We manually transcribed a part of the 2017 spoken data to  train and evaluate the ASR system. 2018 spoken data were used to train and evaluate the grading system. 
Each spoken utterance received a total score from human experts, computed by summing up the scores related to the following $6$ individual indicators: 
 {\bf answer relevance}\COMMENT{(pertinenza della risposta)} (with respect to the question);
{\bf syntactical correctness}\COMMENT{(correttezza competenze formali)} (formal competences, morpho-syntactical correctness); 
{\bf lexical properties}\COMMENT{(propriet\`a lessicali)} (lexical richness and correctness);
{\bf pronunciation}\COMMENT{(pronuncia)}; 
{\bf fluency}\COMMENT{(fluenza)}; 
{\bf communicative skills}\COMMENT{(efficacia comunicativa)} (communicative efficacy and argumentative abilities).

 
Since every utterance was scored by only one expert, it was not possible to evaluate any kind of agreement among experts. However, according to \cite{ramana2017} and \cite{nicolao2015}, inter-rater human correlation varies between around $0.6$ and $0.9$, depending on the type of proficiency test. In this work, correlation between an automatic rater and an expert one is between $0.53$ and $0.61$, indicating a  good performance of the proposed system. For future evaluations more experts are expected to provide independent scoring on the same data sets, so a more precise evaluation will be possible. At present it is not possible to publicly distribute the data.

\vspace{-0.2cm}

\begin{table}[hbt]
\caption{Spoken data collected during the 2017 and 2018 evaluation campaigns. Column ``\#Q'' indicates the total number of different (written) questions presented to the pupils. }
\label{tab:dataspeech}
\centering
\begin{tabular}{ l c r r r r }
\\
\hline
 Year &Lang & \#Pupils   & \#Utterances    & Duration  & \#Q  \\ \hline
 2017 & ENG &  511 &  4112 & 16:25:45 &  24 \\ 
 2017 & GER &  478 &  3739 & 15:33:06 &  23 \\ 
 2018 & ENG & 2331 & 15770 & 93:14:53 &  24 \\ 
 2018 & GER & 2057 & 13658 & 95:54:56 &  23 \\ \hline
 \hline 
\end{tabular}
\end{table}
\vspace{-0.45cm}
\subsection{Manual transcription of spoken data}
\label{sec:manualtrans}
In order to create adaptation and evaluation sets for ASR,
we manually transcribed part of the 2017 data. 
Guidelines for the manual annotation required a trade-off between transcription accuracy and speed. We defined guidelines, where:
\COMMENT{We defined an initial set of guidelines for the annotation, which were used by 5 researchers to manually transcribe about 20 minutes of audio data. This experience lead to a discussion, from which originated a second set of guidelines, meant to reach a reasonable trade-off between transcription accuracy and speed. The main outcomes we found were:}
\textit{a)} only the main speaker has to be transcribed; presence of other voices (school-mates, teacher) should be reported only with the label ``@voices'',
\textit{b)} presence of whispered speech was found to be significant, so it should be explicitly marked with the label ``()'',
\textit{c)} badly pronounced words have to be marked by a ``\#'' sign (without trying to phonetically transcribe the pronounced sounds), and
\textit{d)} code switched words (i.e.\ speech in a different language from the target language) has to be reported by means of an explicit marker, like in: {\it ``I am 10 years old @it(io ho gi\`a risposto)''}.

\COMMENT{Next, we concatenated utterances to be transcribed into blocks of about 5 minutes each. We noticed that knowing the question and hearing several answers could be of great help in transcribing some poorly pronounced words or phrases. Because of this, each block contained only answers to the same question, explicitly reported at the beginning of the block. 
After that, most}
Most of 2017 data was manually transcribed by students from two Italian linguistic high schools (``Curie'' and ``Scholl'') and double-checked by researchers. Part of the data were independently transcribed by pairs of students 
in order to compute inter-annotator agreement, which is shown in Table~\ref{tab:agreement} in terms of Word Accuracy (WA), using the first transcription as a reference (after removing hesitations and other labels related to background voices and noises, etc.). The low level of agreement reflects the difficulty of the task, although it should be noted that the transcribers themselves were non-native speakers of English/German. ASR results will also be affected by this uncertainty.

\COMMENT{We should reduce this part. Not necessary to mention students, only report inter-annotator agreement. 
We engaged about 30 students from two Italian linguistic high schools (namely ``Curie'' and ``Scholl'') to perform manual transcriptions. 
After a joint training session,
we paired students together. Each pair first transcribed a common block of $5$ minutes. Then, they went through a comparison phase, where each pair of students discussed their choices and agreed on a unique transcription for this data block.  Transcriptions made before the comparison phase were retained to evaluate inter-annotator agreement. 
Apart from this first 5 minute block, each utterance was transcribed by only one transcriber.
}  

\begin{table}[th]
\caption{Inter-annotator agreement between pairs of students, in terms of Word Accuracy. Students transcribed English utterances first and German ones later.}
\label{tab:agreement}
\centering
\begin{tabular}{ c c c c c }
\\
\hline
High   &Language &\#Transcribed &\#Different &Agreement \\
school &         &words         &words       & (WA)     \\
\hline
Curie & English  &  965 & 237 & 75.44\% \\
Curie & German   &  822 & 139 & 83.09\% \\
\hline
Scholl & English & 1370 & 302 & 77.96\% \\
Scholl & German  & 1290 & 226 & 82.48\% \\
\hline
\end{tabular}
\end{table}



For both ASR and grading NN experiments, data from the student populations (2017/2018) were divided by speaker identity into training and evaluation sets, with proportions of $\frac{2}{3}$ and $\frac{1}{3}$, respectively (students across the training and evaluation sets do not overlap). 
\COMMENT{By combining  the three student populations with the two input modalities (written and spoken) we obtain six data sets.  
\rg{We only report written modality, although we use written responses for LM training.}
}
Table~\ref{tab:TestAsrV6} reports data about the spoken data set. The id {\em All} identifies the whole data set, while {\em Clean} defines the subset in which sentences containing background voices, incomprehensible speech and fragment words
were excluded.

\begin{table}[th]
\caption{Statistics about the spoken data sets (2017) used for ASR.}
\label{tab:TestAsrV6}
\centering
\begin{tabular}{ l | c | c c | c c } 
\\
\hline
id             & \# of  & \multicolumn{2}{c|}{duration} & \multicolumn{2}{c}{tokens} \\ 
                  &  utt.  & total    &    avg  & total   &    avg \\ \hline
 Ger Train All    &  1448  & 04:47:45 &  11.92  &  9878   &   6.82 \\  
 Ger Train Clean  &   589  & 01:37:59 &   9.98  &  2317   &   3.93 \\ \hline
 Eng Train All    &  2301  & 09:03:30 &  14.17  & 26090   &  11.34 \\
 Eng Train Clean  &   916  & 02:45:42 &  10.85  &  6249   &   6.82 \\ \hline \hline
 Ger Test All     &   671  & 02:19:10 &  12.44  &  5334   &   7.95 \\  
 Ger Test Clean   &   260  & 00:43:25 &  10.02  &  1163   &   4.47 \\ \hline
 Eng Test All     &  1142  & 04:29:43 &  14.17  & 13244   &  11.60 \\  
 Eng Test Clean   &   423  & 01:17:02 &  10.93  &  3404   &   8.05 \\ \hline
\end{tabular}
\end{table}

\vspace{-0.4cm}

\section{ASR system}
\label{sec:ASR}

\subsection{Acoustic model}

The recognition of non-native speech, especially in the framework of multilingual speech recognition, is a well-investigated problem. Past research has tried to model the pronunciation errors of non-native speakers 
\cite{bouselmi2006} 
both by using non-native pronunciation lexicons 
\cite{Wang2003b,Oh2006,strik2009} 
or by adapting acoustic models with either native data and non native data \cite{duan2017,li2016,lee2015,das2015}.

For the recognition of non-native speech, we demonstrated in \cite{matassoni2018} the effectiveness of adapting a multilingual deep neural network (DNN) trained on recordings of native speakers to children between 9 and 14 years old.

In this work we have adopted a more advanced neural network architecture for the multilingual acoustic model, using a time-delay neural network (TDNN) and the popular lattice-free maximum mutual information training (LF-MMI) \cite{povey2016}. The corresponding recipe features i-vector computation, data augmentation via speed perturbation, data clean up and the mentioned MMI training.
Acoustic model training is performed on the following datasets:

$\bullet$ {\bf GerTrainAll} and {\bf EngTrainAll} in-domain sets;

$\bullet$  {\bf Child}, collected in the past in our labs, formed by speech  of Italian children speaking: Italian ({\em ChildIt} subset \cite{gerosa2007} ), English ({\em ChildEn}  \cite{batliner2005}) and German ({\em ChildDe}). 
The children were instructed to read words or sentences in Italian, English or German, respectively;
it contains 28,128 utterances in total, from 249 child speakers between 6 and 13 years old, comprising  44.5 hours of speech.

$\bullet$ {\bf ISLE}, the Interactive Spoken Language Education corpus \cite{menzel2000isle}, consisting of 7,714 read utterances from 23 Italian and 23 German adult learners of English, comprising 9.5 hours of speech. 


\subsection{Language models for ASR}
\label{subsec:lm4asr}
To train effective LMs in this particular domain, we needed sentences capable of representing the simple language spoken by the learners. For each language, we created three sets of text data.
The first included simple texts, collected by grabbing data from Internet pages containing foreign language courses or sample texts (about 113K words for English, 12K words for German). 
The second included training data from the written responses on the written portion of the proficiency tests, acquired during the 2016, 2017  and 2018 evaluation campaigns (see Table~\ref{tab:dataspeech}) (about 393K words for English, 247K words for German). This data underwent a cleaning phase, in which we corrected the most common errors (i.e.\ {\em ai em $\rightarrow$ I am}, {\em becouse  $\rightarrow$ because}, {\em seher $\rightarrow$ sehr}, {\em br\"uder $\rightarrow$ bruder}...) and removed unknown words.
The third included the manual transcriptions of the 2017 spoken data set (see Section~\ref{sec:manualtrans}) (26K words for English, 10K words for German). In this case, we cleaned the data by deleting all markers indicating presence of extraneous phenomena, except for hesitations, which were retained.


This small amount of data (about 532K words for English and 269K words for German in total) was used to train two 3-gram LMs.

\subsection{ASR performance}
\label{subsec:asrperformance}

Table \ref{tab:wers_2017} reports word error rates (WERs) obtained on the 2017 test set (see Table~\ref{tab:TestAsrV6}) using acoustic models trained on both out-of-domain data and in-domain data, which contributes to modeling spontaneous speech and spurious speech phenomena (laughing, coughing, \ldots).

A common phonetic lexicon,  defined in terms of the units of the International Phonetic Alphabet (IPA), is shared across the three languages (Italian, German and English) to allow the usage of a single acoustic model. 
The Table also reports results achieved on a clean subset of the test corpus, which was created by removing sentences with unreliable transcriptions, and spurious acoustic events. 

\begin{table}[h]
\caption{WER results on 2017 spoken test sets.
}
\label{tab:wers_2017}
\centering
\begin{tabular}{ c c c c }
\\
 \hline
 GerTestAll & GerTestClean & EngTestAll & EngTestClean   \\ \hline 
42.6 & 37.5 & 35.9 & 32.6  \\ 
\hline
\end{tabular}
\end{table}




\vspace{-0.4cm}

\section{Proficiency estimation}
\label{sec:classification}

\vspace{-0.1cm}
\subsection{Scoring system}
\label{subsec:profcla}

The classification task consists of predicting the scores assigned by human experts to the spoken answers. For each utterance, 6 scores are given according to the proficiency indicators (described in section \ref{sec:speechdata}). The scores are in the set $\{0,1,2\}$ where 0 indicates not-passed, 1 almost-passed and 2 passed. 

\COMMENT{The examples reported in Table~\ref{tab:writtensamples} refer to the case of 6 indicators. In the Table the total score is computed as the arithmetic sum of the individual scores. However, the proposed scoring system does not compute the total score by simply summing the automatically estimated individual scores, as the experts did; instead, it also automatically estimates the total score from the data.}
\COMMENT{YES but what characterizes each NN is just the training set, not the architecture; also, I don't see any significant difference between the single-indicator NN and the global-score NN}

For estimating each individual score, we employed feed forward NNs, using the corresponding scores assigned to each sentence by the experts as targets. In all cases the score provided by the system corresponds to the index of the output node with the maximum value.
All NNs are trained using the features described below, and are characterized by three layers of dimension equal to the feature size; they use $ReLU$ as activation function, and SGD and $AdaGrad$ for the optimizer. The learning rate is set to 0.05.

Questions associated with the answers are hierarchically clustered according to: {\it a)} the language, {\it b)} the proficiency level (i.e.\ A1, A2 and B1) and {\it c)} two sessions, $S_1$ containing  common questions of proficiency tests (e.g.\ {\em how old are you?}, {\em where do you come from?}, etc)  and $S_2$ containing specific questions (e.g.\ {\em describe your visit to Venice}, etc), and {\it d)} a question identifier.  NNs used for classification follow this subdivision.




\COMMENT{We decided to approach the automatic scoring task as a classification task, instead of using regression models, as is often done in many works on automated L2 classification, in order to mimic as closely as possible the task of the human experts, who, as previously seen, have to choose within a finite set of scores. }


\COMMENT{
\begin{figure}[t]
  \centering
  \includegraphics[width=\linewidth]{figure.pdf}
  \caption{Schematic diagram of speech production.}
  \label{fig:speech_production}
\end{figure}
}

\vspace{-0.1cm}
\subsection{Classification features}
\label{sec:features}

The features used to score the indicators are derived both from the automatic transcriptions of the spoken answers of the students and from the speech signal.

To extract features from the transcriptions we use a set of LMs trained over different types of text data, namely, out-of-domain general texts (for
English, transcriptions of TED talks \cite{falavi2017} - around 3 million words; for German, news - around 1 million words), and in-domain texts containing the ``best'' written/spoken data, collected during the previously mentioned evaluation campaigns carried out in the years 2016, 2017 and 2018. Best data are selected among those that, in the training part, got the highest scores by the experts. 
To compute feature vectors we use $20$ language models, obtained 
by computing {\em four} $n$-gram LMs (estimated by varying the the size of $n$-grams, from 1 to 4) 
over {\em five} different training data sets of decreasing size, as follows: \textit{a)} out-of-domain text data, \textit{b)} all in-domain data, \textit{c)} in-domain data that share the same CEFR proficiency level, \textit{d)} in-domain data that share the same session identifier, and \textit{e)} in-domain data that share the same question identifier. In this way, we assume we know, for each test sentence to score, the language, the intended proficiency level, and the session/question identifiers. 

For each test sentence  formed by $N_W$ words, of which $N_{OOV}$ are out-of-vocabulary (OOV), 
we compute the following $5$ features using each LM (taking inspiration for features from the works of \cite{sakaguchi2015,srihari2007,zechner2009,oh2017,evanini2017}): 

{\em a)} $\frac{\log(P)}{N_W}$, that is, the average log-probability of the sentence, 

{\em b)} $\frac{\log(P_{OOV})}{N_{OOV}}$, that is, the  average contribution of OOV words to the log-probability of the sentence, 

{\em c)} $\frac{\log(P) - log(P_{OOV})}{N_W}$, that is, the average log-difference between the two above probabilities,

{\em d)} $N_W - N_{bo}$, where $N_{bo}$ is the number of back-offs applied by the LM to the input sentence (this difference is related to the frequency of n-grams in the sentence that have also been observed in the training set),

{\em e)} $N_{OOV}$, the number of OOVs in the sentence. 

Note that if  word counts $N_W$ or $N_{OOV}$ are equal to zero (i.e. both $P$ and $P_{OOV}$ are not defined), the corresponding average log-probabilities are replaced by -1.

In this way we compute $5\times 4 \times5=100$ features for each input sentence (five data sets $\times$ four n-gram levels $\times$ five features). To this set we add $11$ more transcription-based features, i.e.: the total number $N_W$ of words in the sentence, the number of content words, the number of OOVs and the percentage of OOVs wrt a reference lexicon, the numbers of words used in Italian, in English and in German, the number of words that had to be corrected by our in-house spelling corrector adapted to this task, the number $BoW$ (Bag of Words) of content words that match the most frequent ones of the ``best'' answers in the training data, $BoW$ divided by all words in the sentence and $BoW$ divided by all the content words in the sentence. This results in a vector of 111 features.

Finally, we use the acoustic model outputs to generate 5 additional pronunciation-based features. For this we used our best non-native model, and two additional native-language acoustic models (one trained on adult English speech from TED talks, and one trained on adult German speech from the BAS corpus). The acoustic model outputs include an alignment of the acoustic frames to phone states and a likelihood of being in that phone state given the acoustic features. We removed frames aligned to silence/background noise, and then generated the following features: \textit{a)} the length of the utterance in number of acoustic frames, \textit{b)} the number of silence frames in the utterance output by our best ASR system, \textit{c)} a confidence score based on the sum of the likelihoods of each context independent phone class, similar to the work by \cite{franco1997automatic}, but averaged over all states in the utterance for that phone class and normalized over the unique phones in the utterance, \textit{d)} the edit distance between the phonetic outputs from the native ASR system and the non-native ASR, similar to \cite{zechner2006towards}, \textit{e)} the difference between the confidence scores from the native and non-native ASR system, similar to \cite{landini2017}.
Therefore, in total, when making use of all features, we represent the student's answer with a vector of dimensionality $116$.

\COMMENT{
For feature selection we adopted a simple but rather fast algorithm that computes the correlation of all given features to the indicators of all levels via Spearman's r (with generally very low p values). Then the 20 most related features for each the spoken and the written data set were selected, by averaging their occurrences over all levels. The analysis had also showed that if a feature was well correlated to one indicator, it was also correlated to the other related ones of the same level/question, thus the averaging was done on the total indicators. We did however not compute inter-feature correlation to exclude features that are highly correlated, as often suggested in the literature, but this could be done for future improvement.\\}

\vspace{-0.4cm}
\section{Classification results and conclusions}
\label{sec:experiments}

\COMMENT{This section has to be revised deeply. We do not report the results  using total score and we report the results of individual indicators with AA and kappa.}

For measuring the performance, we consider three metrics: 
Correct Classification ({\em CC}), linear Weighted Kappa ({\em WK}),
Correlation ({\em Corr}) between the expert's scores and the predicted ones.
For all the three metrics, the value of $1.0$ corresponds to perfect classification; completely wrong classification
is $0.0$ for CC and WK, $-1.0$ for Corr.
For each indicator, we used the training data to train a feedforward NN by grouping sentences sharing language, proficiency level and  session. 
Average classification results are reported in Table~\ref{tab:spokenres}.

\vspace{-0.2cm}

\begin{table}[htb]
\centering
\caption{Average classification results on 2018 data, spoken, reported both on training and test data, given in terms of Correct Classification ({\em CC}), Weighted Kappa ({\em WK}), Correlation ({\em Corr}).}
\label{tab:spokenres}
\begin{tabular}{|l|ccc||ccc|} 
\multicolumn{7}{}{} \\
\hline
Dataset & \multicolumn{3}{c||}{English} &  \multicolumn{3}{c|}{German} \\
        & {\em CC}  & {\em WK}  & {\em Corr} & {\em CC}  & {\em WK}  & {\em Corr}     \\
\hline
  2018 train &  0.712 & 0.840 & 0.684 & 0.763 & 0.866 & 0.763 \\ 
  2018 test  &  0.596 & 0.775 & 0.532 & 0.667 & 0.822 & 0.613 \\ 
\hline
\end{tabular}
\end{table}

Looking at the results in  Table~\ref{tab:spokenres}, the performance in terms of all reported metrics ($CC$, $WK$ and $Corr$) is good, showing that the automatically assigned scores are not far from the manual ones assigned by human experts. The low difference between the performance on training and corresponding test sets indicate that the models do not overfit the data. More importantly, the values of the achieved correlation coefficients resemble those reported in \cite{ramana2017}, related to human rater correlation, on a conversational task which is, in terms of difficulty for L2 learners, similar to some of the tasks analyzed in this paper.

Future work will address the usage of both features selection and  regression models for proficiency classification. Also the investigation of an extended set of features, partially inspired by the hybrid set of features proposed in \cite{falavi2017} for ASR quality estimation will be carried out.



\COMMENT{
now it is a footnote, first page
\section{Acknowledgements}

This work has been partially funded by IPRASE\footnote{http://www.iprase.tn.it} under the project ``TLT -  Trentino Language Testing 2018''. We thank ISIT\footnote{http://www.isit.tn.it} for having provided the reference scores. 
}

\COMMENT{
\begin{figure}[htb]

\begin{minipage}[b]{1.0\linewidth}
  \centering
  \centerline{\includegraphics[width=8.5cm]{image1}}
  \centerline{(a) Result 1}\medskip
\end{minipage}
\begin{minipage}[b]{.48\linewidth}
  \centering
  \centerline{\includegraphics[width=4.0cm]{image3}}
  \centerline{(b) Results 3}\medskip
\end{minipage}
\hfill
\begin{minipage}[b]{0.48\linewidth}
  \centering
  \centerline{\includegraphics[width=4.0cm]{image4}}
  \centerline{(c) Result 4}\medskip
\end{minipage}
\caption{Example of placing a figure with experimental results.}
\label{fig:res}
\end{figure}
}



\label{sec:refs}

\begin{small}
\bibliographystyle{IEEEbib}
\bibliography{paper}
\end{small}
\end{document}